\documentclass[letterpaper, 10 pt, conference]{ieeeconf}  

\IEEEoverridecommandlockouts                              

\usepackage{graphicx}
\usepackage{amsmath}
\usepackage{amssymb}
\usepackage{booktabs}
\usepackage{subcaption}
\usepackage{algorithm}
\usepackage{algorithmic}
\usepackage{setspace}
\usepackage{bbold}
\usepackage{color}
\usepackage[table,xcdraw]{xcolor}
\usepackage{xcolor}
\usepackage{cite}

\title{\LARGE \bf
DiffuPose: Monocular 3D Human Pose Estimation\\via Denoising Diffusion Probabilistic Model  
}

\author{Jeongjun Choi*$^{1,2}$, Dongseok Shim*$^{1}$ and H. Jin Kim$^{1,2}$
\thanks{This research was supported by Unmanned Vehicles Core Technology Research and Development Program through the National Research  Foundation of Korea(NRF) and Unmanned Vehicle Advanced Research Center(UVARC) funded by the Ministry of Science and ICT, the Republic of Korea(NRF-2020M3C1C1A01086411) Also, this research was supported by Institute of Information \&
communications Technology Planning \& Evaluation (IITP)
grant funded by the Korea government(MSIT) [NO.2021-
0-01343, Artificial Intelligence Graduate School Program
(Seoul National University)].}
\thanks{*Equal contribution}
\thanks{$^{1}$The authors are with the Artificial Intelligence Institute of Seoul National University (AIIS) and $^{2}$Automation and Systems Research Institute (ASRI). 
        {\tt\small \{lojol2327, tlaehdtjr01, hjinkim\}@snu.ac.kr}}}%

\begin{document}

\maketitle
\thispagestyle{empty}
\pagestyle{empty}

\begin{abstract}

 Thanks to the development of 2D keypoint detectors, monocular 3D human pose estimation (HPE) via 2D-to-3D uplifting approaches have achieved remarkable improvements.
    Still, monocular 3D HPE is a challenging problem due to the inherent depth ambiguities and occlusions.
    To handle this problem, many previous works exploit temporal information to mitigate such difficulties.
    However, there are many real-world applications where frame sequences are not accessible. 
    This paper focuses on reconstructing a 3D pose from a single 2D keypoint detection.
    Rather than exploiting temporal information, we alleviate the depth ambiguity by generating multiple 3D pose candidates which can be mapped to an identical 2D keypoint.
    We build a novel diffusion-based framework to effectively sample diverse 3D poses from an off-the-shelf 2D detector.
    By considering the correlation between human joints by replacing the conventional denoising U-Net with graph convolutional network, our approach accomplishes further performance improvements.
    We evaluate our method on the widely adopted Human3.6M and HumanEva-I datasets.
    Comprehensive experiments are conducted to prove the efficacy of the proposed method, and they confirm that our model outperforms state-of-the-art multi-hypothesis 3D HPE methods.

\end{abstract}
\section{INTRODUCTION}

Applicability in various fields including human-computer interaction, robotics, sports, and healthcare has driven a significant development in understanding human behavior and estimating an accurate pose \cite{clever20183d, martinez2020residual, zimmermann20183d, reily2020simultaneous}.
For example, accurately detected human pose allows the robot to predict human motions and assist them properly \cite{gui2018teaching, erickson2022characterizing}.
However, estimating human pose from a monocular setting is non-trivial.
Various obstacles such as diverse clothing, background, occlusion, and illumination change make the problem difficult, causing inaccurate pose estimation.
Furthermore, in contrast to 2D keypoint detection task, estimating human pose in 3D space with a single view suffers from inherent depth ambiguity.
The difficulty of obtaining precise 3D annotation data in diverse settings is another big problem.

Fortunately, aided by the development of 2D pose detectors, recent studies \cite{Liu_2020_CVPR, chen2020anatomy, wang2020motion} have achieved impressive progress using 2D-to-3D uplifting pipeline.
They extract the 3D human pose from the output of off-the-shelf 2D keypoint detectors.
However, potentially many 3D poses can be mapped to an identical 2D pose due to depth ambiguity and occlusion. To overcome these innate limitations, many approaches \cite{pavllo20193d,zheng20213d,li2022exploiting} exploit temporal information to estimate a single 3D human pose from consecutive frames (\textit{i.e., many-to-one}).
Unfortunately, the aforementioned methods are inappropriate for real-world applications since they require a sequence of images to produce the current target pose.
On the other hand, others \cite{martinez2017simple, zhao2019semantic, wehrbein2021probabilistic} attempt to produce 3D output from a single frame (\textit{i.e., one-to-one}).
Since it is difficult to estimate an accurate 3D pose from a single frame due to the depth ambiguity, some approaches \cite{jahangiri2017generating,sharma2019monocular,li2020weakly,wehrbein2021probabilistic,li2022mhformer} generate multiple stochastic 3D outputs from a single 2D input.
They handle depth ambiguity as an inverse problem producing multiple feasible solutions.
Our work falls into this category.

\begin{figure}[t]
    \centering
    \includegraphics[width=0.48 \textwidth]{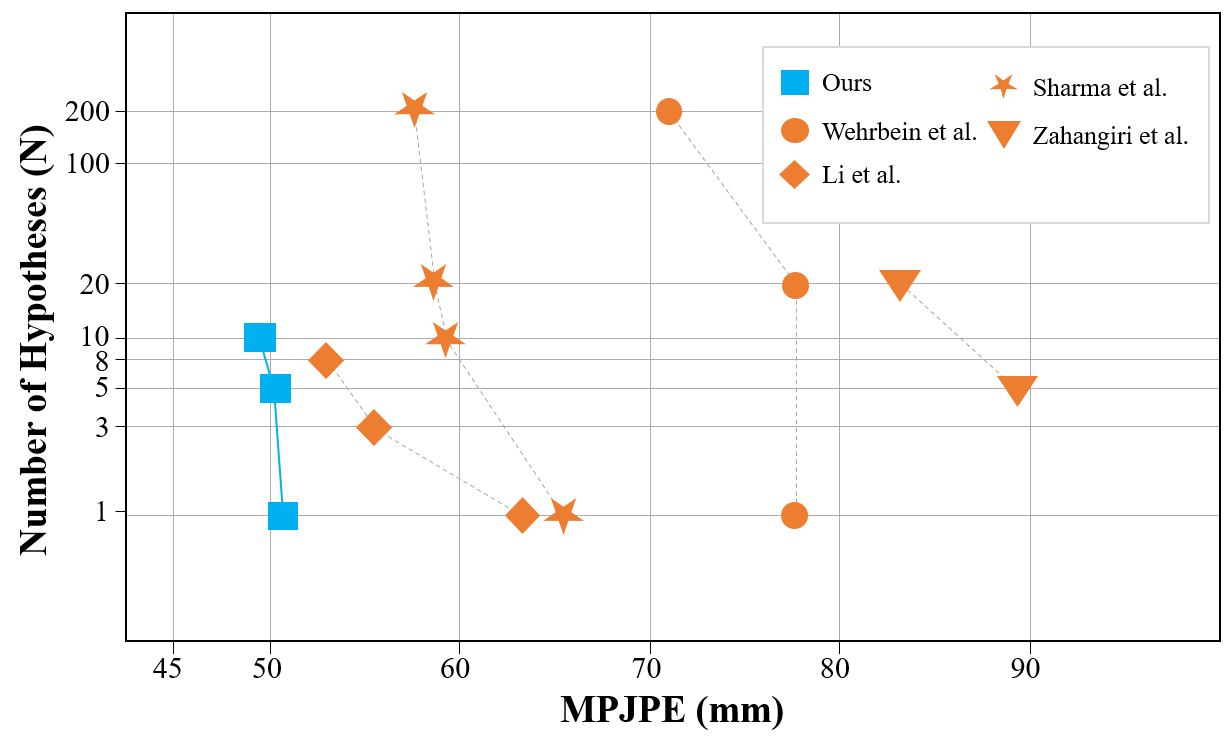}
    \caption{Comparison of Mean Per Joint Position Error (MPJPE) with state-of-the-art multi-hypothesis methods on Human3.6M given a single 2D detection (lower is better). The number of hypotheses is given in the log scale. Our model generates a more plausible set of 3D poses than other state-of-the-art methods, with a smaller number of hypotheses.}
    \vspace{-20pt}
    \label{fig:1}
\end{figure}
\label{sec:intro}
\begin{figure*}[t]
        \centering
    \includegraphics[width=1.0\textwidth]{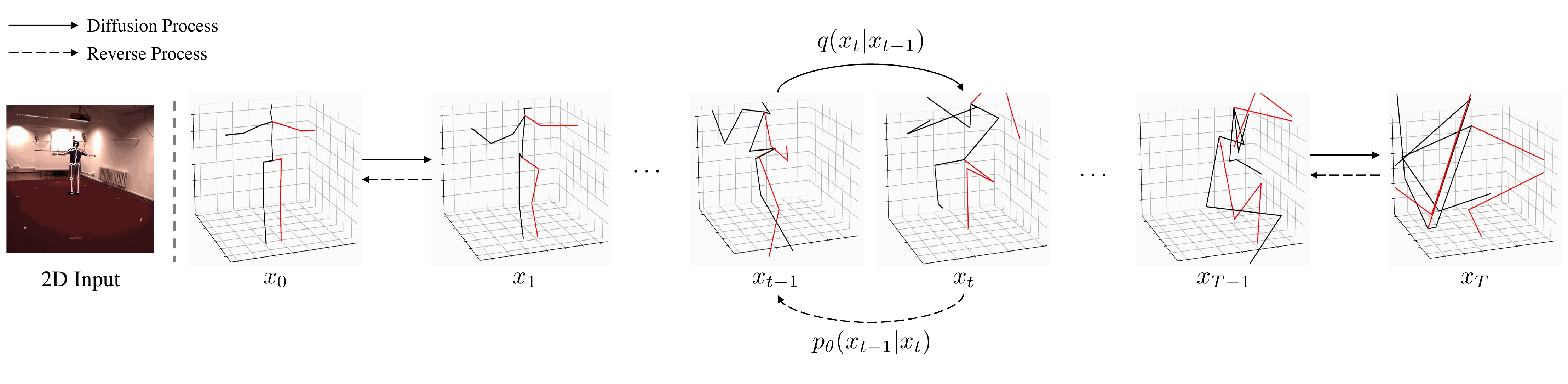}
    \caption{Visualization of the diffusion process for 3D human pose. First, a random noise $x_T$ is sampled from the Gaussian distribution. Then, it is gradually denoised through the reverse process guided by 2D keypoint detection.}
    \label{fig:2}
    \vspace{-10pt}
\end{figure*}

In this work, we propose to produce multiple feasible solutions via denoising diffusion probabilistic model (DDPM) \cite{ho2020denoising}.
DDPM is one of the likelihood-based models which have emerged as the new state-of-the-art in image generation tasks.
By learning how to model the original data distribution in the training phase, diffusion models generate samples with a denoising (backward) process by gradually removing infinitesimal noise from signals.
To the best of our knowledge, this is the first study to exploit DDPM for human pose estimation, in the category of 2D-to-3D lifting method.

Our DiffuPose learns how to reconstruct 3D position of each joint from Gaussian noise with a given 2D keypoint detection obtained from the image in a similar manner to the diffusion models mentioned above.
In a deterministic forward process, the noised 3D feature $x_t$ is sampled from original 3D pose $x_0$ by adding random noise $\epsilon$ from Gaussian distribution, conditioned with diffusion timestep $t$.
Then, the denoising network predicts the added noise $\epsilon_\theta$ so that the denoised 3D feature can be acquired, minimizing the reweighted variational lower bound \cite{ho2020denoising}.
With the predicted 3D feature, the posterior distribution $x_{t-1}$ can be obtained with $x_t$ and $x_0$ in a deterministic form so that the whole denoising step can be learned.
While the depth information can be lost as the 3D feature is mapped to 2D \cite{wehrbein2021probabilistic}, better hypotheses can be extracted by preserving the dimension of 3D features in the input and output of the network.

We address the ambiguous inverse problem of estimating 3D human pose by exploiting the stochastic nature of DDPM.
A number of feasible 3D poses can be generated through the backward process by repeated sampling from Gaussian noise with a single 2D keypoint detection.
For the denoising function in DDPM, we utilize the graph convolutional network of \cite{zou2021modulated} to model the correlation of each joint explicitly.
Experimental results show that our approach can generate more plausible solutions for the inverse problem with a smaller number of samples than previous arts.

We evaluate our framework on the two representative datasets, Human3.6M \cite{ionescu2013human3} and HumanEva-I \cite{sigal2010humaneva}.
Comprehensive experiments show that our method can generate plausible 3D hypotheses.
Furthermore, our method outperforms state-of-the-art multi-hypotheses methods.
Our contribution can be summarized as follows:

\begin{itemize}
    \item We propose a novel 3D human pose estimation network, DiffuPose, which generates multi-hypothesis outputs to alleviate the innate monocular depth ambiguity.
    \item We adopt the graph convolutional network as the denoising function of diffusion models so that the network can explicitly learn the connectivity between the human joints in the 3-dimensional space.
    \item Compared with state-of-the-art multi-hypothesis 3D HPE methods, our method achieves competitive results in Human3.6M dataset.
    
\end{itemize}
\section{RELATED WORK}
\noindent\textbf{2D-to-3D Lifting HPE.}
2D-to-3D lifting approaches can be decomposed into two stages.
They first localize 2D keypoints from image, and then extract them into 3D space.
These methods leverage the excellent performance of 2D pose detectors \cite{cao2017realtime, chen2018cascaded, sun2019deep}.
Estimating 3D pose with 2D detection alleviates the data scarcity problem of 3D HPE since 2D detectors can be trained with both indoor and outdoor data.
Furthermore, these methods can reduce the bias towards indoor scenes, increasing generalization capacity of 3D pose estimation.
Some of 2D-to-3D uplifting approaches exploit temporal information \cite{pavllo20193d, zheng20213d, li2022mhformer} from videos instead of estimating 3D human pose from a single image.
These approaches can improve performance with temporal information, but they are not suitable for real-time applications because they need a long sequence to predict the 3D pose of a central frame.
On the other hand, Martinez \textit{et al.} \cite{martinez2017simple} proposes a network with simple yet effective structure that regresses 3D joints from a single frame.
Mehta \textit{et al.} \cite{mehta2017monocular} proposes fully feedforward convolutional neural network based approach with parent relationships in joint kinematics.
Although our study belongs to this one-to-one category, our model is able to generate \textit{a number of} potential 3D poses given a single 2D detection, whereas the above-mentioned studies only generate \textit{a single} 3D prediction which might be sub-optimal. 

\noindent\textbf{Multi-Hypothesis Methods.}
3D human pose estimation with monocular image is an ill-posed problem in that just regressing a single solution is unlikely to be optimal.

Earlier studies \cite{sminchisescu2003kinematic, simo2012single} point out the depth ambiguity problem of single-view 3D pose estimation, and utilize heuristic methods to generate multiple 3D poses.
In recent, Jahangiri \textit{et al.} \cite{jahangiri2017generating} proposes to generate valid 3D hypotheses consistent with 2D keypoint detection using anatomical constraints such as joint angle and bone length.
Li \textit{et al.} \cite{li2019generating} combines a mixture density network \cite{bishop1994mixture} with the inverse problem of monocular HPE to generate plausible 3D poses.
Oikarinen \textit{et al.} \cite{oikarinen2021graphmdn} improves the mixture density network-based method \cite{li2019generating} with the strength of the semantic graph neural network \cite{zhao2019semantic}.
Li \textit{et al.} \cite{li2022mhformer} proposes a multi-hypothesis transformer to predict a 3D pose of a central frame but it requires a long input sequence of 2D keypoints as additional temporal information.
While the previous works require that the number of samples be set in advance, it can be determined differently for every inference in the following works, including our model.
Sharma \textit{et al.} \cite{sharma2019monocular} employs a conditional variational autoencoder to sample 3D-pose candidates and then aggregates those candidates with ordinal relations predicted from a deep convolutional neural network.
Wehrbein \textit{et al.} \cite{wehrbein2021probabilistic} models the posterior distribution of 3D poses, with normalizing flow by explicitly incorporating the uncertainty of 2D keypoint detector.

Unfortunately, the previous works either need additional temporal information \cite{li2022mhformer}, or perform poorly without \textit{Oracle} supervision which requires 3D ground-truth poses for hypotheses selection \cite{sharma2019monocular, wehrbein2021probabilistic}.
On the other hand, our method shows competitive results with a delicately built diffusion model without extra frames or the hypothesis selection guided by 3D ground-truth poses.

\noindent\textbf{Denoising Diffusion Probabilistic Models.}
Recently, Denoising Diffusion Probabilistic Model (DDPM) \cite{ho2020denoising} achieves the exceptional performance  in density estimation \cite{kingma2021variational} and image synthesis \cite{rombach2022high, dhariwal2021diffusion, song2020denoising}.
Song \textit{et al.} DDIM \cite{song2020denoising} improves the sampling speed of DDPM with deterministic mapping from latent to image by setting the noise value to zero.
SR3 \cite{saharia2022image} proposes conditional DDPM for better super resolution which also requires \textit{one-to-many} mapping to learn the super resolution space.
Latent diffusion \cite{rombach2022high} (LDM) points out the low inference speed and high training costs of conventional diffusion models and brings the denoising process into the compressed latent space with a much lower dimension.

As DDPM and its variants prove their superiority in synthesizing high-quality images or modeling data distribution, we take such advantage of DDPM for 3D human pose estimation.
Due to the inherent depth ambiguity, it is necessary to model the data distribution of 3D pose given a single 2D pose.
The problem with leveraging DDPM in human pose estimation is that the huge success of DDPM attributes to the high inductive bias for image data by implementing backbone architecture with U-Net \cite{ ho2020denoising, dhariwal2021diffusion}.
When it comes to human pose estimation, we need to handle graph-like joints and edges, not image-like 2D data.
As the human pose has tight semantic connectivity between adjacent joints, we adopt graph convolutional neural network (GCN) \cite{kipf2016semi} so that the model explicitly learns the relation between joints.
\section{METHODOLOGY}
\begin{figure*}[t]
    \centering
    \includegraphics[width=1.0\textwidth]{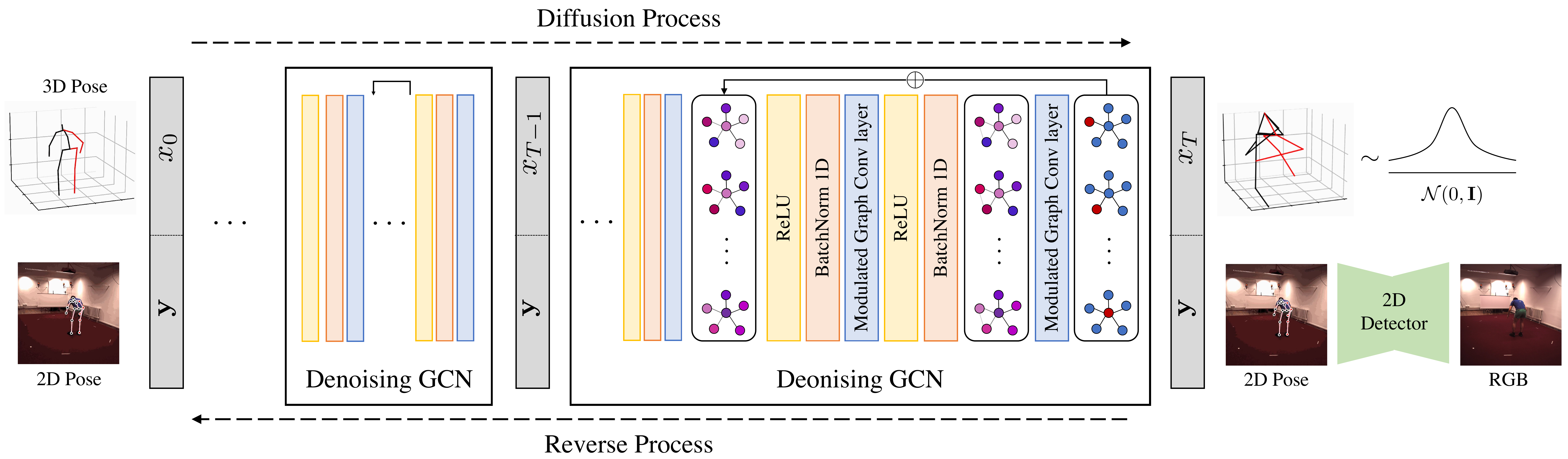}
    \caption{Overview of DiffuPose. 
    A randomly sampled graph with joints $x_{T}$ is randomly sampled from Gaussian distribution and fed to the denoising Graph Convolutional Network (CGN) along with the 2D pose detection results.
    After multiple denoising steps, DiffuPose generates stochastic 3D pose outputs which correspond to the 2D pose input.
    }
    \vspace{-10pt}
    \label{fig:3}
\end{figure*}

In this section, we detail about our proposed DiffuPose.
The entire structure of our proposed network is depicted in Figure \ref{fig:3}.
\subsection{Preliminary}
Denoising Diffusion Probabilistic Model (DDPM) \cite{ho2020denoising} is a kind of generative model that utilizes a parameterized Markov chain with variational inference to synthesize the data in the complex distribution from a simple data distribution.
Specifically, from the target data $x_{0}$, the forward process $q$ gradually adds infinitesimal Gaussian noise, $\epsilon$, with variance of $\beta_{t}\in[0,1]$ at time $t$ through $x_{1}$ to $x_{T}$ qs the following:
\begin{equation}
    \begin{split}
        q(x_{1}, \cdots, x_{T}|x_{0}) &:= \prod_{t=1}^{T}q({x_{t}|x_{t-1}})\\
        q(x_{t}|x_{t-1}) &:= \mathcal{N}(x_{t};\sqrt{1-\beta_{t}}x_{t-1}, \beta_{t}\mathbf{I})
    \end{split}
    \label{eq1}
\end{equation}
where $T$ indicates the total diffusion step and $\mathbf{I}$ denotes identity matrix.

From Equation (\ref{eq1}), we can rewrite the forward process to directly sample $x_t$ conditioned only on $x_0$ as
\begin{equation}
        q(x_t|x_0) = \mathcal{N}(x_t;\sqrt{\Bar{\alpha_t}}x_0, (1-\Bar{\alpha_t})\mathbf{I}),
    \label{eq2}
\end{equation}
where $\alpha:=1-\beta_t$ and $\bar{\alpha}_t:=\prod_{s=1}^t\alpha_s$.

By estimating the posterior $q(x_{t-1}|x_{t})$, which is called the reverse process, one can figure out the target data $x_{0}$ or its distribution explicitly.
With the Bayes theorem, the posterior $q(x_{t-1}|x_{t}, x_{0})$ can be formulated as
\begin{equation}
\begin{split}
\tilde{\beta}_{t} &:= \frac{1-\Bar{\alpha}_{t-1}}{1-\Bar{\alpha}_{t}}\beta_{t}\\
\tilde{\mu}_{t}(x_{t}, x_{0}) &:= \frac{\sqrt{\Bar{\alpha}_{t-1}}\beta_{t}}{1-\Bar{\alpha}_{t}}x_{0} + \frac{\sqrt{\alpha_{t}}(1-\Bar{\alpha}_{t-1})}{1-\Bar{\alpha}_{t}}x_{t}\\
&q(x_{t-1}|x_{t}, x_{0}) = \mathcal{N}(\tilde{\mu}_{t}(x_{t}, t), \tilde{\beta}_{t}\mathbf{I})
\end{split}
\label{eq3}
\end{equation}

Unfortunately, as shown in Equation (\ref{eq3}), we need to know $x_{0}$ beforehand to measure the posterior.
DDPM, therefore, does not directly calculate the posterior but rather estimates its value with neural network $p_{\theta}$,
\begin{equation}
    p_{\theta}(x_{t-1}|x_{t}) \simeq q(x_{t-1}|x_{t},x_{0}).
\end{equation}

Among many different ways to estimate $p_{\theta}(x_{t-1}|x_{t})$, DDPM alternatively estimates the infinitesimal noise between consecutive timesteps, $\epsilon_{\theta}(x_{t}, t)$. Finally, the loss function for DDPM can be formulated in a reweighted variational lower bound form:
\begin{equation}
    \mathcal{L} = \mathbb{E}_{t,x_{0}, \epsilon}[||\epsilon - \epsilon_{\theta}(x_{t},t)||^{2}].
\end{equation}

\subsection{Graph Convolutional Network}
Unlike the previous DDPM \cite{ho2020denoising} and its variants \cite{song2020denoising, rombach2022high, dhariwal2021diffusion} which employ the U-Net \cite{ronneberger2015u} based backbone for synthesizing image-like 2-dimensional data, 
we adopt graph convolutional network (GCN) \cite{kipf2016semi} as a denoising function of diffusion network for graph-like 3-dimensional human pose data with node and edge.

GCN consists of an affinity matrix $\mathbf{A}\in\mathbb{R}^{N\times N}$ which represents the connectivity between neighboring nodes with binary entries, $a_{i,j} \in \{0,1\}$, and a shared feature embedding matrix $\mathbf{W}\in \mathbb{R}^{d\times d^{\prime}}$ which maps the feature of each node from $d$ to $d^{\prime}$ dimensional latent space.
A single graph convolutional layer aggregates the feature from graph-like input as follows:
\begin{equation}
    \mathbf{H}^{\prime} = \sigma(\mathbf{W}\mathbf{H}\mathbf{\tilde{A}}),
\end{equation}
where $\mathbf{H}$ indicates the input of GCN layer whose column is the per-node feature, and $\mathbf{H}^{\prime}$ represents the output of GCN layer.
$\sigma$ denotes a non-linear activation function such as ReLU, and $\mathbf{\tilde{A}}$ is a symmetrically normalized affinity matrix \cite{kipf2016semi}.

We do not directly adopt vanilla GCN, but rather, employ a modulation strategy for human pose estimation \cite{zou2021modulated} so that the entries of the affinity matrix are no longer limited to the binary value with the learnable masks, $\mathbf{P}$ and $\mathbf{Q}$.
The feature embedding matrix $\mathbf{W}^{\prime}_{j}$ is joint-specifically learned to improve 3D human pose estimation,
\begin{equation}
\begin{split}
    \mathbf{A}_{\mathrm{mod}} &= \mathbf{A} \odot \mathbf{P} + \mathbf{Q},\\
    \mathbf{H}^{\prime} &= \sigma(\mathbf{W}^{\prime}\mathbf{H}\mathbf{\tilde{A}}_{\mathrm{mod}}),    
\end{split}
\end{equation}
where $\odot$ indicates element-wise multiplication and $\mathbf{Q}$ is implemented with the average of a matrix and its transpose $(\mathbf{Q} + \mathbf{Q^{\top}})/2$, to force the affinity matrix to be symmetric for undirected graph structure.

Also, for the efficiency of computation in GCN, we utilize the weight modulation \cite{zou2021modulated} because joint-specific feature mapping requires much more computation compared to vanilla GCN.
To alleviate such inefficiency, we do not implement joint-wise matrix multiplication.
Instead, we map the node-specific features with the shared weight and unshare them by joint-specific light matrix multiplication \cite{zou2021modulated}.
By doing so, we can take advantage of rich representation from joint-specific feature mapping without heavy computation.

By utilizing GCN instead of naive U-Net based diffusion architecture, our model can learn the conditional distribution of 3D pose better since GCN explicitly considers the connectivity between human joints.


\subsection{Diffusion Pipeline for 3D Pose Estimation}
We propose to incorporate a 2D detection into denoising network with 3D joints in order to generate a desired 3D pose.
First, we sample denoising step $t \sim U(0,T)$ to generate corrupted 3D joints $\mathbf{x_t}\in\mathbb{R}^{J\times3}$ from the ground-truth 3D joints $\mathbf{x_0}$ with Gaussian noise $\epsilon$ according to Equation (\ref{eq2}).
Then, the output of off-the-shelf 2D human pose detector $\mathbf{y}\in\mathbb{R}^{J\times2}$ is spatially concatenated with $\mathbf{x_t}$.
In the diffusion process, the correlation between 2D and 3D joints is implicitly learned for each denoising step as the fused feature go through the GCN-based network.
To reflect the deoising step $t$, the timestep-residual block is added between modulated GCN blocks.
We follow \cite{dhariwal2021diffusion} for the time embedding so that the network is effectively guided by $t$.
The output of the network predicts $\epsilon$ directly, so we optimize the network for 3D noise with
\begin{equation}
    \mathcal{L}_1 = \mathbb{E}_{t,\mathbf{x}_{0}, \epsilon \sim \mathcal{N}(\mathbf{0}, \mathbf{I})}[||\epsilon - \epsilon_{\theta}(\mathbf{x}_{t},t, \mathbf{y})||].
\end{equation}
Additionally, to make the 3D joints generation process more compliant with 2D detection, we train the network to reconstruct y in a supervised manner:
\begin{equation}
    \mathcal{L}_2 = \mathbb{E}_{t,\mathbf{y}}[||\mathbf{y} - \hat{\mathbf{y}}_{\theta}(\mathbf{x}_{t},t, \mathbf{y})||],
\end{equation}
where $y$ is the output of the 2D detector and $\hat{y}$ is the reconstructed detection.
Note that the network is optimized jointly for $\epsilon_\theta$ and $\mathbf{\hat{y}}$.
Also, we penalize each joint with different weights for joint-specific attention.
The final loss function can be written as:
\begin{equation}
    \mathcal{L} = \mathcal{L}_1 + \lambda\mathcal{L}_2.
\end{equation}
By setting the loss term in this way, 2D condition can be imposed without any computationally intensive operations like cross-attention.

In the reverse process, $\mathbf{x}_T$ is extracted from the isotropic Gaussian.
Given a 2D detection $\mathbf{y}$, the random noise $\mathbf{x}_T$ is gradually denoised through the learned diffusion network.
Fig \ref{fig:2} illustrates the gradual denoising process, by which a 3D pose is reconstructed from a random noise conditioned with a single 2D keypoint detection.

Additionally, as we aim to reconstruct a single 3D pose, we adopt the cosine noise scheduler suggested in \cite{dhariwal2021diffusion}:
\begin{equation}
    \bar{\alpha}_t = \frac{f(t)}{f(0)}, \quad f(t) = \cos(\frac{t/T+s}{1+s}\cdot\frac{\pi}{2})^2
\end{equation}
which is proven to be useful for reconstructing data of small size.
It helps to add noise more slowly than linear schedule \cite{ho2020denoising}, achieving better log-likelihoods.
Through the above process, our network can generate a plausible set of 3D poses which are closely linked to a single 2D keypoint by repeated sampling.
\begin{table*}[t]
    \centering
    \resizebox{\textwidth}{!}{
    \begin{tabular}{l|ccccccccccccccc|c}
        \toprule
        MPJPE (mm) & Dir. & Disc. & Eat & Greet & Phone & Photo & Pose & Pur. & Sit & SitD. & Smoke & Wait & WalkD. & Walk & WalkT. & \textbf{Avg.} \\
        \hline
        Martinez \textit{et al.} \cite{martinez2017simple} & 51.8 & 56.2 & 58.1 & 59.0 & 69.5 & 78.4 & 55.2 & 58.1 & 74.0 & 94.6 & 62.3 & 59.1 & 65.1 & 49.5 & 52.4 & 62.9 \\
        Pavllo \textit{et al.} \cite{pavllo20193d}  (T=1) ($\dagger$) & 47.1 &50.6& 49.0& 51.8& 53.6& 61.4& 49.4& 47.4& 59.3& 67.4& 52.4& 49.5& 55.3& 39.5& 42.7& 51.8\\
        Pavllo \textit{et al.} \cite{pavllo20193d} (T=243) ($\dagger$) &45.2 &46.7& 43.3& 45.6& 48.1& 55.1 &44.6 &44.3& 57.3& 65.8& 47.1 &44.0&49.0& 32.8& 33.9& 46.8\\
        Li \textit{et al.} \cite{li2022mhformer} (T=351) ($\dagger$) &39.2& 43.1 &40.1 &40.9 &44.9& 51.2& 40.6& 41.3& 53.5& 60.3& 43.7& 41.1& 43.8& 29.8& 30.6 &43.0\\
        Cai \textit{et al.} \cite{cai2019exploiting} (T=1) ($\dagger$) & 46.5& 48.8& 47.6& 50.9& 52.9& 61.3 &48.3& 45.8& 59.2& 64.4& 51.2& 48.4& 53.5& 39.2& 41.2& 50.6\\
        
        Lin \textit{et al.} \cite{lin2021end} & -& -& - &-& -& -& -& - &- &- &- &- &-& -& -& 54.0\\
        Xu \textit{et al.} \cite{xu2021graph} & 45.2& 49.9& 47.5& 50.9& 54.9& 66.1& 48.5& 46.3& 59.7& 71.5& 51.4& 48.6& 53.9& 39.9& 44.1& 51.9\\
        \hline
        Jahangiri \textit{et al.} \cite{jahangiri2017generating} (N=5)  & 82.9 & 77.5 & 81.6 & 85.2 & 90.9 & 80.5 & 78.8 & 109.3 & 138.7 & 97.8 & 90.1 & 86.4 & 77.9 & 85.5 & 81.5 & 89.2 \\
        Jahangiri \textit{et al.} \cite{jahangiri2017generating} (N=20)   & 77.1 & 71.2 & 75.4 & 79.0 & 84.7 & 74.9 & 72.4 & 102.2 & 131.5 & 85.9 & 84.5 & 80.4 & 71.6 & 78.4 & 74.9 & 82.9 \\
        Sharma \textit{et al.} \cite{sharma2019monocular} (N=1)*  & 49.3 & 61.8 & 61.2 & 67.1 & 62.9 & 69.9 & 57.1 & 69.6 & 85.9 & 81.3 & 58.8 & 67.5 & 66.6 & 51.0 & 58.1 & 64.5 \\
        Sharma \textit{et al.} \cite{sharma2019monocular} (N=20)* & 44.2 & 56.1 & 54.3 & 58.2 & 56.0 & 61.2 & 50.5 & 63.1 & 83.3 & 75.5 & 52.5 & 61.6 & 59.4 & 45.2 & 52.5 & 58.4 \\
        Sharma \textit{et al.} \cite{sharma2019monocular} (N=200)*  & 43.9 & 55.5 & 52.6 & 56.8 & 55.2 & 60.0 & 50.0 & 61.5 & 83.1 & 74.8 & 51.7 & 60.2 & 57.7 & 44.6 & 52.0 & 57.6 \\
        Li \textit{et al.} (N=10) \cite{li2020weakly} &62.0 &69.7& 64.3& 73.6& 75.1& 84.8& 68.7& 75.0& 81.2& 104.3& 70.2& 72.0& 75.0& 67.0& 69.0& 73.9 \\   
        Li \textit{et al.} (N=5) \cite{li2019generating} & 43.8 & \textbf{48.6} & 49.1 & \textbf{49.8} & 57.6 & 61.5 & \textbf{45.9} & 48.3 & 62.0 & 73.4 & 54.8 & 50.6 & 56.0 & 43.4 & 45.5 & 52.7\\
        Oikarinen \textit{et al.} \cite{oikarinen2021graphmdn} (N=5)  & 49.9 & 54.9 & 55.2 & 56.0 & 62.1 & 73.2 & 51.6 & 53.2 & 69.0 & 88.2 & 58.9 & 55.8 & 61.0 & 48.6 & 50.1 & 59.2 \\
        Wehrbein \textit{et al.} \cite{wehrbein2021probabilistic} ($z_0$) (N=1) & 52.4 & 60.2 & 57.8 & 57.4 & 65.7 & 74.1 & 56.2 & 59.1 & 69.3 & 78.0 & 61.2 & 63.7 & 67.0 & 50.0 & 54.9 & 61.8 \\
        Wehrbein \textit{et al.} \cite{wehrbein2021probabilistic} (N=1)* & 67.5 & 74.7 & 70.9 & 73.4 & 78.5 & 87.9 & 70.1 & 74.3 & 81.0 & 93.1 & 75.7 & 79.3 & 81.3 & 70.4 & 66.7& 76.3\\
        Wehrbein \textit{et al.} \cite{wehrbein2021probabilistic} (N=20)* & 68.9 & 76.4 & 70.3 & 74.3 & 79.1 & 89.0 & 72.2 & 75.5 & 82.2 & 93.6 & 75.9 & 79.6 & 82.1 & 71.8 & 67.4 & 77.2\\
        Wehrbein \textit{et al.} \cite{wehrbein2021probabilistic} (N=200)* & 63.2 & 71.1 & 65.7 & 68.7 & 74.1 & 84.4 & 67.5 & 70.0 & 77.1 & 88.1 & 70.5 & 74.4 & 76.1 & 66.0 & 61.2 & 71.9 \\
        \hline
        Ours (Baseline) & 50.3 & 54.9 & 58.7 & 58.6 & 60.5 & 65.8 &52.6  &51.6  &66.1  &79.1  & 57.6 & 56.9 & 60.1 & 47.6 & 50.6 & 58.1 \\
        Ours (N=1) & 44.3 & 51.6 & 46.3 & 51.1 & 50.3 & 54.3 &49.4 & 45.9 & 57.7 & 71.6 & 48.6 & 49.1 & 52.1 & 44.0 & 44.4 & 50.7\\
        Ours (N=10) & \textbf{43.4} & 50.7 & \textbf{45.4} & 50.2 & \textbf{49.6} & \textbf{53.4} & 48.6 & \textbf{45.0} & \textbf{56.9} & \textbf{70.7} & \textbf{47.8} & \textbf{48.2}  & \textbf{51.3} & \textbf{43.1}  & \textbf{43.4} & \textbf{49.4} \\
        \midrule\midrule
        P-MPJPE (mm) & Dir. & Disc. & Eat & Greet & Phone & Photo & Pose & Pur. & Sit & SitD. & Smoke & Wait & WalkD. & Walk & WalkT. & \textbf{Avg.} \\
        \hline
        Martinez \textit{et al.} \cite{martinez2017simple} & 39.5& 43.2& 46.4& 47.0& 51.0& 56.0& 41.4& 40.6& 56.5& 69.4& 49.2& 45.0& 49.5& 38.0& 43.1 &47.7\\
        Pavllo \textit{et al.} \cite{pavllo20193d} (T=1) ($\dagger$)  &36.0 &38.7& 38.0& 41.7& 40.1& 45.9& 37.1& 35.4& 46.8& 53.4& 41.4& 36.9& 43.1& 30.3& 34.8& 40.0\\
        Pavllo \textit{et al.} \cite{pavllo20193d} (T=243) ($\dagger$) &34.1 &36.1& 34.4 &37.2& 36.4& 42.2& 34.4& 33.6 &45.0& 52.5& 37.4& 33.8& 37.8& 25.6& 27.3& 36.5\\
        Li \textit{et al.} \cite{li2022mhformer} (T=351) ($\dagger$) &31.5& 34.9& 32.8 &33.6& 35.3& 39.6& 32.0 &32.2 &43.5& 48.7& 36.4& 32.6& 34.3& 23.9 &25.1 &34.4\\
        Cai \textit{et al.} \cite{cai2019exploiting} (T=1)($\dagger$)  &36.8& 38.7& 38.2& 41.7& 40.7& 46.8& 37.9& 35.6& 47.6 &51.7& 41.3& 36.8 &42.7& 31.0 &34.7 &40.2
        \\\hline
        Sharma \textit{et al.} \cite{sharma2019monocular} (N=1)  & 35.3& 35.9& 45.8& 42.0& 40.9& 52.6& 36.9& 35.8& 43.5& 51.9& 44.3& 38.8& 45.5& 29.4& 34.3& 40.9  \\
        Li \textit{et al.} \cite{li2020weakly} (N=10)  &38.5& 41.7& 39.6& 45.2& 45.8& 46.5& 37.8& 42.7& 52.4& 62.9& 45.3& 40.9& 45.3& 38.6& 38.4& 44.3 \\        
        Li \textit{et al.} \cite{li2019generating} (N=5)  & \textbf{35.5}& \textbf{39.8}& 41.3 &42.3& 46.0& 48.9& \textbf{36.9}& 37.3& 51.0& 60.6& 44.9& 40.2& 44.1& 33.1& 36.9& 42.6 \\
        Oikarinen \textit{et al.} \cite{oikarinen2021graphmdn} (N=5)  & 38.5 & 42.6 & 44.1 & 44.9 & 48.1 & 53.3 & 39.0 & 39.5 & 54.9 & 66.2 & 47.0 & 42.2 & 46.8 & 36.8 & 39.8 & 45.6 \\
        Wehrbein \textit{et al.} \cite{wehrbein2021probabilistic} ($z_{0}$) (N=1)  & 37.8& 41.7& 42.1& 41.8& 46.5& 50.2& 38.0& 39.2& 51.7& 61.8& 45.4& 42.6& 45.7& 33.7& 38.5& 43.8 \\
        Wehrbein \textit{et al.} \cite{wehrbein2021probabilistic} (N=1)* & 47.4 & 50.8 & 50.4 & 51.8 & 54.8 & 58.4 & 46.8 & 49.6 & 58.3 & 71.4 & 54.2 & 51.9 & 54.3 & 47.3 & 44.5 & 52.8 \\
        Wehrbein \textit{et al.} \cite{wehrbein2021probabilistic} (N=20)* & 47.3 & 51.1 & 49.6 & 51.5 & 54.7 & 58.2 & 47.1 & 48.9 & 59.4 & 71.1 & 54.0 &52.2& 54.4 & 48.5 & 44.8 & 52.9 \\
        Wehrbein \textit{et al.} \cite{wehrbein2021probabilistic} (N=200)* & 42.8 & 46.6 & 45.9 & 47.1 & 50.4 & 54.4 & 43.0 & 44.3 & 55.2 & 66.5 & 49.6 & 47.5 & 49.9 & 43.6 & 39.4 & 48.4 \\
        \hline
        Ours (Baseline) & 42.1 & 44.8 & 47.2 & 48.8 & 47.6 & 51.8 & 40.7 & 40.8 & 54.6 & 66.6 & 47.2 & 44.8 & 50.6 & 38.4 & 42.5 & 47.2\\
        Ours (N=1) & 36.7 & 41.1 & 37.6 & 42.2 & 40.5 & 44.1 & 37.8 & 36.3 & 47.0 & 60.5 & 39.8 & 38.9 & 42.7 & 33.7 & 35.1 & 40.9\\
        Ours (N=10) & 35.9 & 40.3 & \textbf{36.7}&\textbf{41.4} & \textbf{39.8} & \textbf{43.4} & 37.1 & \textbf{35.5} & \textbf{46.2} & \textbf{59.7} & \textbf{39.9} & \textbf{38.0}& \textbf{41.9} & \textbf{32.9}  & \textbf{34.2}& \textbf{39.9} \\
        \bottomrule
    \end{tabular}
    }
    \caption{Detailed quantitative results of MPJPE in millimeters on Human3.6M under Protocol 1 (no rigid alignment) and Protocol 2 (rigid alignment). \textbf{Top}: results under Protocol 1 (MPJPE); \textbf{Bottom}: results under Protocol 2 (P-MPJPE). Results with (*) are computed from the officially released code from \cite{sharma2019monocular} and \cite{wehrbein2021probabilistic}. ($\dagger$) indicates using temporal information, and T denotes the number of input frames. N denotes the number of samples estimated by respective approaches.}
    \label{tab:table1}
    \vspace{-10pt}
\end{table*}

\section{EXPERIMENTS}

\subsection{Datasets and Evaluation Metrics}
\textbf{Human3.6M} \cite{ionescu2013human3} is currently the largest indoor dataset for 3D human pose estimation.
There exist 11 professional actors performing 15 activities such as \textit{Sitting, Discussing, Greeting}.
Videos were recorded from 4 synchronized cameras in 50Hz, and accurate 2D and 3D joint annotations and camera parameters are provided.
Following the previous policy \cite{pavllo20193d, chen2020anatomy, Liu_2020_CVPR, wang2020motion},  the model is trained on 5 subjects (S1, S5, S6, S7, S8) and tested on 2 subjects (S9, S11). 
For evaluation metric, we report the two most commonly used protocols for Human3.6M dataset.
As protocol 1, MPJPE (Mean Per Joint Position Error) is computed as the mean Euclidean distance between the estimated and ground-truth 3D joints in millimeters.
Protocol 2 is P-MPJPE (Procrustes MPJPE), which is computed the same as MPJPE after rigidly aligning the estimated 3D joints to the ground-truth.

\textbf{HumanEva-I} \cite{sigal2010humaneva} is also a popular 3D human pose dataset, which is much smaller than Human3.6M.
It contains 7 video sequences captured from MoCap.
4 actors perform 6 different actions such as \textit{Walk, Jog, Gesture}, etc.
We use protocol 2 (P-MPJPE) as error metric.
Following \cite{martinez2017simple,pavllo20193d}, we divided the data into train/validation sets with \textit{Walk, Jog, Box} action, and the validation set is used for testing.

\begin{table*}[t]
    \centering
    \resizebox{\textwidth}{!}
    {
    \begin{tabular}{l|ccccccccccccccc|c}
        \toprule
        MPJPE (mm) & Dir. & Disc. & Eat & Greet & Phone & Photo & Pose & Pur. & Sit & SitD. & Smoke & Wait & WalkD. & Walk & WalkT. & \textbf{Avg.} \\
        \hline
        Zhou \textit{et al.} \cite{zhou2019hemlets} (+)& 34.4 & 42.4 & 36.6 & 42.1 & 38.2 & 39.8 & 34.7 & 40.2 & 45.6 & 60.8 & 39.0 & 42.6 & 42.0 & 29.8 & 31.7 & 39.9 \\ 
        Ci \textit{et al.} \cite{ci2019optimizing} (+)(*)& 36.3 & 38.8 & 29.7 & 37.8 & 34.6 & 42.5 & 39.8 & 32.5 & 36.2 & 39.5 & 34.4 & 38.4 & 38.2 & 31.3 & 34.2 & 36.3 \\
        \hline
        Pavllo \textit{et al.} \cite{pavllo20193d} (T=9) ($\dagger$)  & 37.0 & 40.7 & 35.2 & 37.4 & 38.4 & 44.2 & 42.3 & 37.1 & 46.5 & 48.8 & 38.9 & 40.1 & 38.5 & 29.9 & 32.6 &39.2\\ 
        Zheng \textit{et al.} \cite{zheng20213d} (T=9) ($\dagger$)& 49.2 & 49.7 & 38.7 & 42.7 & 40.0 & 40.9 & 50.7 & 42.2 & 47.0 & 46.1 & 43.4 & 46.7 & 39.8 & 36.4 & 38.0 & 43.5 \\
        Zheng \textit{et al.} \cite{zheng20213d} (T=81) ($\dagger$) & 30.0 & 33.6 & 29.9&  31.0& 30.2& 33.3 &34.8& 31.4& 37.8 &38.6& 31.7 &31.5 &29.0 &23.3& 23.1& 31.3\\
        Li \textit{et al.} \cite{li2022mhformer} (T=351) ($\dagger$) &27.7 &32.1& 29.1& 28.9 &30.0& 33.9 &33.0 &31.2& 37.0& 39.3 &30.0 &31.0& 29.4& 22.2 &23.0& 30.5\\

       \midrule
           Martinez \textit{et al.} \cite{martinez2017simple}& 45.2&46.7&43.3&45.6&48.1 & 55.1 & 44.6 & 44.3 & 57.3 & 65.8 & 47.1 & 44.0 & 49.0 &  32.8& 33.9&46.8 \\ 
           Zhao \textit{et al.} \cite{zhao2019semantic}& 37.8&49.4&37.6&40.9&45.1 & 41.4 & 40.1 & 48.3 & 50.1 & 42.2 & 53.5 & 44.3 & 40.5 &  47.3& 39.0&43.8 \\ 
               Liu \textit{et al.} \cite{liu2020comprehensive}& 36.8&40.3&33.0&36.3&37.5 & 45.0 & 39.7 & 34.9 & 40.3 & 47.7 & 37.4 & 38.5 & 38.6 & 29.6& 32.0&37.8 \\ 
                   Xu \textit{et al.} \cite{xu2021graph}& 35.8&38.1&31.0&35.3&35.8 & 43.2 & 37.3 & 31.7 & 38.4 & 45.5 & 35.4 & 36.7 & 36.8 & 27.9& 30.7&35.8 \\ 
        \midrule
        Ours (N=10) & 39.9 & 42.5 & 31.2 & 35.9 & 37.5 & 39.5 & 42.4 & 32.6 & 38.8 & 42.4 & 37.0  &39.0 & 36.2& 30.7 &33.8 & 37.3 \\
        \bottomrule
    \end{tabular}
    }
    \caption{Detailed quantitative results of MPJPE in millimeters on Human3.6M under Protocol 1 (no rigid alignment). Ground truth 2D keypoints (poses) are used as input. (+), (*), and ($\dagger$)  denote extra data from MPII \cite{agarwal2005recovering}, pose scaling during both training and testing, and temporal information respectively.}
    \label{tab:gt}
    \vspace{-10pt}
\end{table*}

\subsection{Implementation Details}
The proposed network is implemented with a public deep learning platform, PyTorch \cite{paszke2019pytorch}.
We trained and tested our model using a single NVIDIA RTX A5000 GPU.
We adopted the Adam \cite{kingma2014adam} optimizer with initial learning rate 4e-5, with shrink factor 0.995.
We apply horizontal flip augmentation as in \cite{chen2020anatomy, Liu_2020_CVPR, pavllo20193d} during both training and testing phases.
We trained the model for 200 epochs with batch size 1024 for Human3.6M \cite{ionescu2013human3} and 1000 epochs with batch size 128 for HumanEva-I \cite{sigal2010humaneva}.
We set the diffusion timestep 100 since such small number of diffusion step is enough for a single 3D pose.
For the cosine noise scheduler, we set the offset $s$ 0.008.
We tested for both linear and cosine schedules to validate the effectiveness of the cosine scheduler in low-dimensional data.
The result is illustrated in section 4.4.
For the 2D keypoint input, we use the 2D detection from \cite{sun2019deep} and ground-truth for Human3.6M, and 2D joints provided by the dataset for HumanEva-I.

\begin{table}[t]
    \centering
    \resizebox{0.45\textwidth}{!}
    {
    \begin{tabular}{c|c c c|c c c|c c c}
    \toprule
          & & Walk & & & Jog & & & Box & \\\hline
         N& S1 & S2 & S3 & S1 & S2 & S3 & S1 & S2 & S3\\
    \midrule
         1 &  13.6& 11.6 & 16.7 & 18.7 & 12.7 &12.4 &  18.4 & 23.6 & 21.3\\
         5 & 12.7& 10.6& 16.0 & 18.2&11.9 &11.3 & 17.7 & 23.1 & 20.8\\
         10 & 12.6 &10.5 &15.9 &18.1 &11.8 &11.2 &17.6 &23.0 & 20.7 \\
         30 & 12.5 & 10.4& 15.9& 18.0& 11.7& 11.1& 17.6 & 23.0 & 20.6\\

    \bottomrule
    \end{tabular}
    }
    \caption{Evaluation results for HumanEva-I dataset. N denotes the number of hypotheses. Reported values are the reconstruction error after rigid alignment (P-MPJPE).}
    \label{tab:table5}
\end{table}
\begin{table}[t]
    \centering
    \resizebox{0.4\textwidth}{!}
    {
    \begin{tabular}{c|c|c}
    \toprule
    &MPJPE (mm) ($\downarrow$)& \# of Params ($\downarrow$)\\
    \midrule
         Baseline (U-Net) & 58.1&2.99 M \\
         Ours (GCN)& 50.7 & 1.45 M\\
    \bottomrule
    \end{tabular}
    }
    \caption{Ablation on denoising function for DDPM.
    Graph Convolutional Network (GCN) produces better performance in MPJPE for human pose estimation with fewer model parameters compared to U-Net which is the de-facto standard for image-based DDPM.
    }
    \label{tab:table2}
    \vspace{-10pt}
\end{table}
\begin{table}[t]
    \centering
    \resizebox{0.35\textwidth}{!}
    {
    \begin{tabular}{c|c|c}
    \toprule
         \# of Samples (N)&Dimension ($d_{m}$) & MPJPE (mm) ($\downarrow$)  \\
    \midrule
         1 & 128 & 53.4 \\
         1 & 256 & 50.7 \\
         1 & 384 & 50.0\\
         \hline
         5 & 128 & 51.7 \\
         5 & 256 & 49.9\\
         5 & 384 & 49.5\\
         \hline
         10 & 128 & 51.4 \\
         10 & 256 & 49.8\\
         10 & 384 & 49.4\\
    
    \bottomrule
    \end{tabular}
    }
    \caption{Evaluation results for Human3.6M dataset for various setting.}
    \label{tab:table4}
    \vspace{-20pt}
\end{table}

\subsection{Comparison with State-of-the-art Methods}
\noindent{\textbf{Results on Human3.6M.}}
We compare our proposed method, DiffuPose, with state-of-the-art multi-hypotheses human pose estimation approaches using a single frame in all the 15 actions and their average error in Table \ref{tab:table1}.
The reported results \textit{do not} leverage the ground truth 3D poses for selecting the best models in the prediction sets, rather evaluating the mean value of the prediction sets or method-specific hypotheses selection strategy without ground truth.
It is because selecting best-hypothesis with reference to the ground truth does not perfectly fit the real world where 3D ground truth poses usually are not available.
Table \ref{tab:table1} shows that our method produces the best result for the average MPJPE of 50.7 mm under Protocol 1  and 40.9 mm under Protocol 2 with the sample size of 1.

For multiple hypotheses setting, we randomly sample 10 different outputs with a single 2D detection result and average them to compare with the ground truth 3D human pose.
Our method also achieves the best results with multiple sampling and performs even better than existing methods with larger number of samples.
Specifically, the performance for estimating 3D human pose is improved approximately 6.4\% (3.3 mm) in Protocol 1 and 6.3\% (2.7mm) in Protocol 2 compared to Li \textit{et al.} \cite{li2019generating}.

We also compare our method with state-of-the-art 3D pose estimation algorithms using 2D ground truth, and the result is reported in Table \ref{tab:gt}.
Compared to the methods that predict a 3D pose from a single-frame without additional training data (i.e., \cite{martinez2017simple, zhao2019semantic, liu2020comprehensive, xu2021graph}), our DiffuPose shows competitive result. 
Our result even outperforms some approaches which leverage temporal information with a sequence of 2D keypoints.

\noindent\textbf{Results on HumanEva-I.}
We conduct an experiment to validate our method on a smaller dataset.
We train our model with 3 subjects (\textit{i.e.}, S1, S2, S3) and evaluate on the validation sequences using all 3 subjects as test data.
The validation sequences are separate from the trainset and do not include any of the subjects used for training. We evaluate our model's performance on the validation sequences using all three subjects (S1, S2, S3) as test data. After rigid alignment, we report the error between the model's predictions and the ground-truth data.
Table \ref{tab:table5} shows the reconstruction error according to the number of hypotheses.
We can find that the error decreases in every subject for all 3 actions as N increases.

\begin{figure}[t]
    \centering
    \includegraphics[width=0.5 \textwidth]{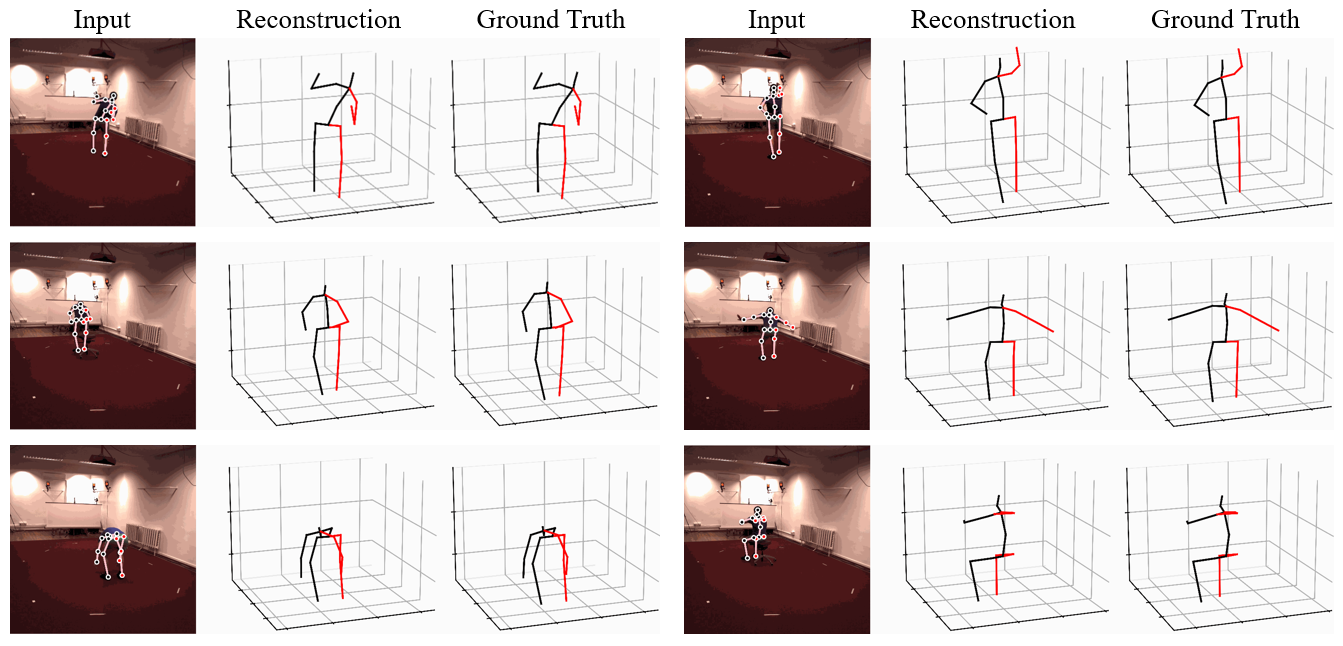}
    \caption{Qualitative results of our DiffuPose with various actions on Human3.6M \cite{ionescu2013human3} test set S11.
    }
    \label{fig:5}
\end{figure}

\noindent\textbf{Qualitative Results.}
The qualitative results of our method are demonstrated in Figure \ref{fig:5}.
For visualization, ground truth 2D input is used.
Our network produces plausible 3D human pose outputs similar to the ground truth pose.

\begin{figure}[t]
    \centering
    \includegraphics[width=0.45 \textwidth]{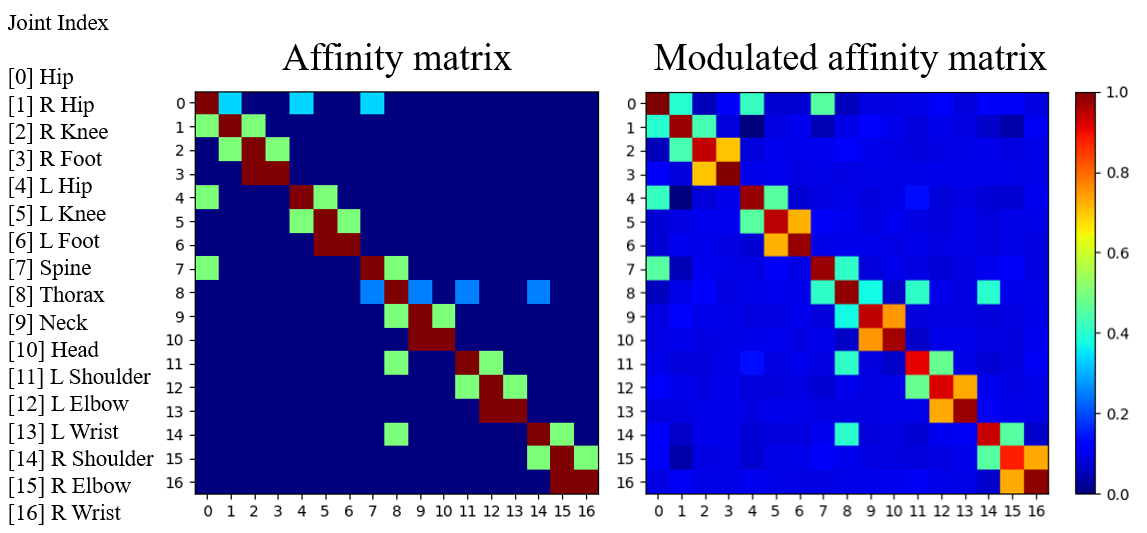}
    \caption{
    Visualization of affinity matrix and modulated affinity matrix.
    Each row and column indicate the index of the joints.
    }
    \label{fig:4}
    \vspace{-10pt}
\end{figure}

\subsection{Ablation Study}
For better understanding of how each component in DiffuPose contributes to the model performance, we perform several ablation studies in this section.

\noindent\textbf{Baseline.}
We explore the effectiveness of the denoising function in DDPM for human pose estimation in Table \ref{tab:table2}.
We observe that leveraging Graph Convolutional Network (GCN) as backbone for DDPM achieves much better performance with less complex model compared to U-Net \cite{ronneberger2015u}, which has been widely adopted as denoising function of DDPM for image input.
GCN improves roughly 7.4 mm (12.6 $\%$) in MPJPE with 52 $\%$ fewer model parameters.

\noindent\textbf{Training / Inference time.}
The overall training procedure takes about 12 hours with one A5000 GPU for 200 epochs in Human3.6M dataset.
For inference time, it takes quite long for diffusion models as reported in \cite{song2020denoising}. However, one can use Denoising Diffusion Implicit Model (DDIM) \cite{song2020denoising} for faster inference, without compensating much accuracy. 
In Table \ref{tab:table6}, we report the number of output frames per second (FPS) according to various number of DDIM steps (totally 100 steps) and corresponding errors. DiffuPose still can achieve competitive performance while easily satisfying the real-time standard.

\noindent\textbf{Parameter Setting Analysis.}
In Table \ref{tab:table4}, we report the effectiveness of the performance according to the setting of model hyperparameters under Protocol 1 with MPJPE.
There are two major hyperparameters of the network, \textit{i.e.} the number of pose hypotheses and the model dimension $d_{m}$, and we evaluate the choice of each configuration.
Based on the results, we choose the combination of $N=10$ and $d_{m}=384$ which produces the best performance with efficient model parameters (3.22M).

\noindent\textbf{Noise Scheduling.}
We conduct an experiment for each noise schedule.
The number of hypotheses and the model dimension is set to 1 and 256, equally for both cases.
With linear noise schedule\cite{ho2020denoising}, we obtain 54.9 mm error for protocol 1.
In the same hyperparameter setting, we obatin 50.7 mm error, which is about 7.7\% lower. We can conclude that the cosine schedule is effective for small data, as in image domain.

\noindent\textbf{Visualization of Affinity Matrix.}
As DiffuPose leverages a learnable affinity matrix for denosing GCN, we qualitatively demonstrate the effectiveness of the modulated affinity matrix compared to the naive binary affinity matrix.
As shown in Figure \ref{fig:4}, the modulated affinity matrix exhibits much broader connection by learning the connectivity between two joints which are physically disconnected but semantically connected such as arms and legs.

\begin{table}[h]
    \centering
    \resizebox{0.45\textwidth}{!}
    {
    \begin{tabular}{c|c|c|c}
    \toprule
        DDIM steps & FPS & MPJPE (mm) (N=1) & MPJPE (mm) (N=10) \\
    \midrule
         2 & 8927 & 56.7 & 50.2\\
         5 & 3820 & 53.8 & 49.8\\
         10 & 1795 & 53.7 & 49.7\\
         20 & 869 & 53.6 & 49.6\\
         100 (DDPM) & 188 & 50.7 & 49.4\\

    \bottomrule
    \end{tabular}
    }
    \caption{Analysis on inference speed. N denotes the number of hypotheses.}
    \vspace{-20pt}
    \label{tab:table6}
\end{table}


\section{CONCLUSIONS}

In this paper, we present a novel framework named DiffuPose for 2D-to-3D uplifting monocular human pose estimation with diffusion model.
Our model gradually reconstructs 3D human pose from the Gaussian noise given a single 2D detection.
By doing so, DiffuPose can address the depth ambiguity problem better.
As a result, our method achieves competitive results in a single-frame monocular setting, outperforming state-of-the-art multi-hypotheses methods in Human3.6M dataset.




\bibliographystyle{IEEEtran}
\bibliography{root.bib}

\begin{thebibliography}{10}
\providecommand{\url}[1]{#1}
\csname url@samestyle\endcsname
\providecommand{\newblock}{\relax}
\providecommand{\bibinfo}[2]{#2}
\providecommand{\BIBentrySTDinterwordspacing}{\spaceskip=0pt\relax}
\providecommand{\BIBentryALTinterwordstretchfactor}{4}
\providecommand{\BIBentryALTinterwordspacing}{\spaceskip=\fontdimen2\font plus
\BIBentryALTinterwordstretchfactor\fontdimen3\font minus
  \fontdimen4\font\relax}
\providecommand{\BIBforeignlanguage}[2]{{%
\expandafter\ifx\csname l@#1\endcsname\relax
\typeout{** WARNING: IEEEtran.bst: No hyphenation pattern has been}%
\typeout{** loaded for the language `#1'. Using the pattern for}%
\typeout{** the default language instead.}%
\else
\language=\csname l@#1\endcsname
\fi
#2}}
\providecommand{\BIBdecl}{\relax}
\BIBdecl

\bibitem{clever20183d}
H.~M. Clever, A.~Kapusta, D.~Park, Z.~Erickson, Y.~Chitalia, and C.~C. Kemp,
  ``3d human pose estimation on a configurable bed from a pressure image,'' in
  \emph{2018 IEEE/RSJ International Conference on Intelligent Robots and
  Systems (IROS)}.\hskip 1em plus 0.5em minus 0.4em\relax IEEE, 2018, pp.
  54--61.

\bibitem{martinez2020residual}
A.~Mart{\'\i}nez-Gonz{\'a}lez, M.~Villamizar, O.~Can{\'e}vet, and J.-M. Odobez,
  ``Residual pose: A decoupled approach for depth-based 3d human pose
  estimation,'' in \emph{2020 IEEE/RSJ International Conference on Intelligent
  Robots and Systems (IROS)}.\hskip 1em plus 0.5em minus 0.4em\relax IEEE,
  2020, pp. 10\,313--10\,318.

\bibitem{zimmermann20183d}
C.~Zimmermann, T.~Welschehold, C.~Dornhege, W.~Burgard, and T.~Brox, ``3d human
  pose estimation in rgbd images for robotic task learning,'' in \emph{2018
  IEEE International Conference on Robotics and Automation (ICRA)}.\hskip 1em
  plus 0.5em minus 0.4em\relax IEEE, 2018, pp. 1986--1992.

\bibitem{reily2020simultaneous}
B.~Reily, Q.~Zhu, C.~Reardon, and H.~Zhang, ``Simultaneous learning from human
  pose and object cues for real-time activity recognition,'' in \emph{2020 IEEE
  international conference on robotics and automation (ICRA)}.\hskip 1em plus
  0.5em minus 0.4em\relax IEEE, 2020, pp. 8006--8012.

\bibitem{gui2018teaching}
L.-Y. Gui, K.~Zhang, Y.-X. Wang, X.~Liang, J.~M. Moura, and M.~Veloso,
  ``Teaching robots to predict human motion,'' in \emph{2018 IEEE/RSJ
  International Conference on Intelligent Robots and Systems (IROS)}.\hskip 1em
  plus 0.5em minus 0.4em\relax IEEE, 2018, pp. 562--567.

\bibitem{erickson2022characterizing}
Z.~Erickson, H.~M. Clever, V.~Gangaram, E.~Xing, G.~Turk, C.~K. Liu, and C.~C.
  Kemp, ``Characterizing multidimensional capacitive servoing for physical
  human--robot interaction,'' \emph{IEEE Transactions on Robotics}, 2022.

\bibitem{Liu_2020_CVPR}
R.~Liu, J.~Shen, H.~Wang, C.~Chen, S.-c. Cheung, and V.~Asari, ``Attention
  mechanism exploits temporal contexts: Real-time 3d human pose
  reconstruction,'' \emph{In CVPR}, 2020.

\bibitem{chen2020anatomy}
T.~Chen, C.~Fang, X.~Shen, Y.~Zhu, Z.~Chen, and J.~Luo, ``Anatomy-aware 3d
  human pose estimation in videos,'' \emph{IEEE Transactions on Circuits and
  Systems for Video Technology}, 2021.

\bibitem{wang2020motion}
J.~Wang, S.~Yan, Y.~Xiong, and D.~Lin, ``Motion guided 3d pose estimation from
  videos,'' 2020.

\bibitem{pavllo20193d}
D.~Pavllo, C.~Feichtenhofer, D.~Grangier, and M.~Auli, ``3d human pose
  estimation in video with temporal convolutions and semi-supervised
  training,'' in \emph{Proceedings of the IEEE/CVF Conference on Computer
  Vision and Pattern Recognition}, 2019, pp. 7753--7762.

\bibitem{zheng20213d}
C.~Zheng, S.~Zhu, M.~Mendieta, T.~Yang, C.~Chen, and Z.~Ding, ``3d human pose
  estimation with spatial and temporal transformers,'' in \emph{Proceedings of
  the IEEE/CVF International Conference on Computer Vision}, 2021, pp.
  11\,656--11\,665.

\bibitem{li2022exploiting}
W.~Li, H.~Liu, R.~Ding, M.~Liu, P.~Wang, and W.~Yang, ``Exploiting temporal
  contexts with strided transformer for 3d human pose estimation,'' \emph{IEEE
  Transactions on Multimedia}, 2022.

\bibitem{martinez2017simple}
J.~Martinez, R.~Hossain, J.~Romero, and J.~J. Little, ``A simple yet effective
  baseline for 3d human pose estimation,'' in \emph{Proceedings of the IEEE
  international conference on computer vision}, 2017, pp. 2640--2649.

\bibitem{zhao2019semantic}
L.~Zhao, X.~Peng, Y.~Tian, M.~Kapadia, and D.~N. Metaxas, ``Semantic graph
  convolutional networks for 3d human pose regression,'' in \emph{Proceedings
  of the IEEE/CVF conference on computer vision and pattern recognition}, 2019,
  pp. 3425--3435.

\bibitem{wehrbein2021probabilistic}
T.~Wehrbein, M.~Rudolph, B.~Rosenhahn, and B.~Wandt, ``Probabilistic monocular
  3d human pose estimation with normalizing flows,'' in \emph{Proceedings of
  the IEEE/CVF international conference on computer vision}, 2021, pp.
  11\,199--11\,208.

\bibitem{jahangiri2017generating}
E.~Jahangiri and A.~L. Yuille, ``Generating multiple diverse hypotheses for
  human 3d pose consistent with 2d joint detections,'' in \emph{Proceedings of
  the IEEE International Conference on Computer Vision Workshops}, 2017, pp.
  805--814.

\bibitem{sharma2019monocular}
S.~Sharma, P.~T. Varigonda, P.~Bindal, A.~Sharma, and A.~Jain, ``Monocular 3d
  human pose estimation by generation and ordinal ranking,'' in
  \emph{Proceedings of the IEEE/CVF international conference on computer
  vision}, 2019, pp. 2325--2334.

\bibitem{li2020weakly}
C.~Li and G.~H. Lee, ``Weakly supervised generative network for multiple 3d
  human pose hypotheses,'' 2020.

\bibitem{li2022mhformer}
W.~Li, H.~Liu, H.~Tang, P.~Wang, and L.~Van~Gool, ``Mhformer: Multi-hypothesis
  transformer for 3d human pose estimation,'' in \emph{Proceedings of the
  IEEE/CVF Conference on Computer Vision and Pattern Recognition}, 2022, pp.
  13\,147--13\,156.

\bibitem{ho2020denoising}
J.~Ho, A.~Jain, and P.~Abbeel, ``Denoising diffusion probabilistic models,''
  \emph{Advances in Neural Information Processing Systems}, vol. volume 33, pp.
  pages 6840--6851, 2020.

\bibitem{zou2021modulated}
Z.~Zou and W.~Tang, ``Modulated graph convolutional network for 3d human pose
  estimation,'' in \emph{Proceedings of the IEEE/CVF International Conference
  on Computer Vision}, 2021, pp. 11\,477--11\,487.

\bibitem{ionescu2013human3}
C.~Ionescu, D.~Papava, V.~Olaru, and C.~Sminchisescu, ``Human3. 6m: Large scale
  datasets and predictive methods for 3d human sensing in natural
  environments,'' \emph{IEEE transactions on pattern analysis and machine
  intelligence}, vol.~36, no.~7, pp. 1325--1339, 2013.

\bibitem{sigal2010humaneva}
L.~Sigal, A.~O. Balan, and M.~J. Black, ``Humaneva: Synchronized video and
  motion capture dataset and baseline algorithm for evaluation of articulated
  human motion,'' \emph{International journal of computer vision}, vol.~87,
  no.~1, pp. 4--27, 2010.

\bibitem{cao2017realtime}
Z.~Cao, T.~Simon, S.-E. Wei, and Y.~Sheikh, ``Realtime multi-person 2d pose
  estimation using part affinity fields,'' in \emph{Proceedings of the IEEE
  conference on computer vision and pattern recognition}, 2017, pp. 7291--7299.

\bibitem{chen2018cascaded}
Y.~Chen, Z.~Wang, Y.~Peng, Z.~Zhang, G.~Yu, and J.~Sun, ``Cascaded pyramid
  network for multi-person pose estimation,'' in \emph{Proceedings of the IEEE
  conference on computer vision and pattern recognition}, 2018, pp. 7103--7112.

\bibitem{sun2019deep}
K.~Sun, B.~Xiao, D.~Liu, and J.~Wang, ``Deep high-resolution representation
  learning for human pose estimation,'' in \emph{Proceedings of the IEEE/CVF
  conference on computer vision and pattern recognition}, 2019, pp. 5693--5703.

\bibitem{mehta2017monocular}
D.~Mehta, H.~Rhodin, D.~Casas, P.~Fua, O.~Sotnychenko, W.~Xu, and C.~Theobalt,
  ``Monocular 3d human pose estimation in the wild using improved cnn
  supervision,'' in \emph{2017 international conference on 3D vision
  (3DV)}.\hskip 1em plus 0.5em minus 0.4em\relax IEEE, 2017, pp. 506--516.

\bibitem{sminchisescu2003kinematic}
C.~Sminchisescu and B.~Triggs, ``Kinematic jump processes for monocular 3d
  human tracking,'' in \emph{2003 IEEE Computer Society Conference on Computer
  Vision and Pattern Recognition, 2003. Proceedings.}, vol.~1.\hskip 1em plus
  0.5em minus 0.4em\relax IEEE, 2003, pp. I--I.

\bibitem{simo2012single}
E.~Simo-Serra, A.~Ramisa, G.~Alenya, C.~Torras, and F.~Moreno-Noguer, ``Single
  image 3d human pose estimation from noisy observations,'' in \emph{2012 IEEE
  Conference on Computer Vision and Pattern Recognition}.\hskip 1em plus 0.5em
  minus 0.4em\relax IEEE, 2012, pp. 2673--2680.

\bibitem{li2019generating}
C.~Li and G.~H. Lee, ``Generating multiple hypotheses for 3d human pose
  estimation with mixture density network,'' in \emph{Proceedings of the
  IEEE/CVF conference on computer vision and pattern recognition}, 2019, pp.
  9887--9895.

\bibitem{bishop1994mixture}
C.~M. Bishop, ``Mixture density networks,'' 1994.

\bibitem{oikarinen2021graphmdn}
T.~Oikarinen, D.~Hannah, and S.~Kazerounian, ``Graphmdn: Leveraging graph
  structure and deep learning to solve inverse problems,'' in \emph{2021
  International Joint Conference on Neural Networks (IJCNN)}.\hskip 1em plus
  0.5em minus 0.4em\relax IEEE, 2021, pp. 1--9.

\bibitem{kingma2021variational}
D.~Kingma, T.~Salimans, B.~Poole, and J.~Ho, ``Variational diffusion models,''
  \emph{Advances in neural information processing systems}, vol.~34, pp.
  21\,696--21\,707, 2021.

\bibitem{rombach2022high}
R.~Rombach, A.~Blattmann, D.~Lorenz, P.~Esser, and B.~Ommer, ``High-resolution
  image synthesis with latent diffusion models,'' in \emph{Proceedings of the
  IEEE/CVF Conference on Computer Vision and Pattern Recognition}, 2022, pp.
  10\,684--10\,695.

\bibitem{dhariwal2021diffusion}
P.~Dhariwal and A.~Nichol, ``Diffusion models beat gans on image synthesis,''
  \emph{Advances in Neural Information Processing Systems}, vol.~34, pp.
  8780--8794, 2021.

\bibitem{song2020denoising}
J.~Song, C.~Meng, and S.~Ermon, ``Denoising diffusion implicit models,'' in
  \emph{International Conference on Learning Representations}, 2020.

\bibitem{saharia2022image}
C.~Saharia, J.~Ho, W.~Chan, T.~Salimans, D.~J. Fleet, and M.~Norouzi, ``Image
  super-resolution via iterative refinement,'' \emph{IEEE Transactions on
  Pattern Analysis and Machine Intelligence}, 2022.

\bibitem{kipf2016semi}
T.~N. Kipf and M.~Welling, ``Semi-supervised classification with graph
  convolutional networks,'' \emph{arXiv preprint arXiv:1609.02907}, 2016.

\bibitem{ronneberger2015u}
O.~Ronneberger, P.~Fischer, and T.~Brox, ``U-net: Convolutional networks for
  biomedical image segmentation,'' in \emph{International Conference on Medical
  image computing and computer-assisted intervention}.\hskip 1em plus 0.5em
  minus 0.4em\relax Springer, 2015, pp. 234--241.

\bibitem{cai2019exploiting}
Y.~Cai, L.~Ge, J.~Liu, J.~Cai, T.-J. Cham, J.~Yuan, and N.~M. Thalmann,
  ``Exploiting spatial-temporal relationships for 3d pose estimation via graph
  convolutional networks,'' in \emph{Proceedings of the IEEE/CVF international
  conference on computer vision}, 2019, pp. 2272--2281.

\bibitem{lin2021end}
K.~Lin, L.~Wang, and Z.~Liu, ``End-to-end human pose and mesh reconstruction
  with transformers,'' in \emph{Proceedings of the IEEE/CVF Conference on
  Computer Vision and Pattern Recognition}, 2021, pp. 1954--1963.

\bibitem{xu2021graph}
T.~Xu and W.~Takano, ``Graph stacked hourglass networks for 3d human pose
  estimation,'' in \emph{Proceedings of the IEEE/CVF conference on computer
  vision and pattern recognition}, 2021, pp. 16\,105--16\,114.

\bibitem{zhou2019hemlets}
K.~Zhou, X.~Han, N.~Jiang, K.~Jia, and J.~Lu, ``Hemlets pose: Learning
  part-centric heatmap triplets for accurate 3d human pose estimation,'' in
  \emph{Proceedings of the IEEE/CVF international conference on computer
  vision}, 2019, pp. 2344--2353.

\bibitem{ci2019optimizing}
H.~Ci, C.~Wang, X.~Ma, and Y.~Wang, ``Optimizing network structure for 3d human
  pose estimation,'' in \emph{Proceedings of the IEEE/CVF international
  conference on computer vision}, 2019, pp. 2262--2271.

\bibitem{liu2020comprehensive}
K.~Liu, R.~Ding, Z.~Zou, L.~Wang, and W.~Tang, ``A comprehensive study of
  weight sharing in graph networks for 3d human pose estimation,'' in
  \emph{European Conference on Computer Vision}.\hskip 1em plus 0.5em minus
  0.4em\relax Springer, 2020, pp. 318--334.

\bibitem{agarwal2005recovering}
A.~Agarwal and B.~Triggs, ``Recovering 3d human pose from monocular images,''
  \emph{IEEE transactions on pattern analysis and machine intelligence},
  vol.~28, no.~1, pp. 44--58, 2005.

\bibitem{paszke2019pytorch}
A.~Paszke, S.~Gross, F.~Massa, A.~Lerer, J.~Bradbury, G.~Chanan, T.~Killeen,
  Z.~Lin, N.~Gimelshein, L.~Antiga \emph{et~al.}, ``Pytorch: An imperative
  style, high-performance deep learning library,'' \emph{Advances in neural
  information processing systems}, vol.~32, 2019.

\bibitem{kingma2014adam}
D.~P. Kingma and J.~Ba, ``Adam: A method for stochastic optimization,''
  \emph{arXiv preprint arXiv:1412.6980}, 2014.

\end{thebibliography}

\end{document}